\documentclass[10pt,twocolumn,letterpaper]{article}

\usepackage{wacv}              %

\usepackage{graphicx}
\usepackage{amsmath}
\usepackage{amssymb}
\usepackage{booktabs}
\usepackage{xcolor}         %

\usepackage{enumitem}
\usepackage[accsupp]{axessibility} %

\usepackage[pagebackref,breaklinks,colorlinks]{hyperref}

\usepackage[capitalize]{cleveref}
\crefname{section}{Sec.}{Secs.}
\Crefname{section}{Section}{Sections}
\Crefname{table}{Table}{Tables}
\crefname{table}{Tab.}{Tabs.}

\def\wacvPaperID{1837} %
\def\confName{WACV}
\def\confYear{2025}

\definecolor{green}{rgb}{0.059, 0.710, 0.216}
\definecolor{orange}{rgb}{1.0, 0.5, 0.31}
\definecolor{purple}{rgb}{0.596, 0, 0.710}

\begin{document}

\title{Good Seed Makes a Good Crop:\\Discovering Secret Seeds in Text-to-Image Diffusion Models}

\author{%
  Katherine Xu\\
  University of Pennsylvania\\
  \and
  Lingzhi Zhang\\
  Adobe Inc.\\
  \and
  Jianbo Shi\\
  University of Pennsylvania\\
}

\twocolumn[{%
\maketitle
\vspace{-26 pt}
\begin{center}
    \centering
    \includegraphics[trim=0in 1.05in 0in 0in, clip,width=\textwidth]{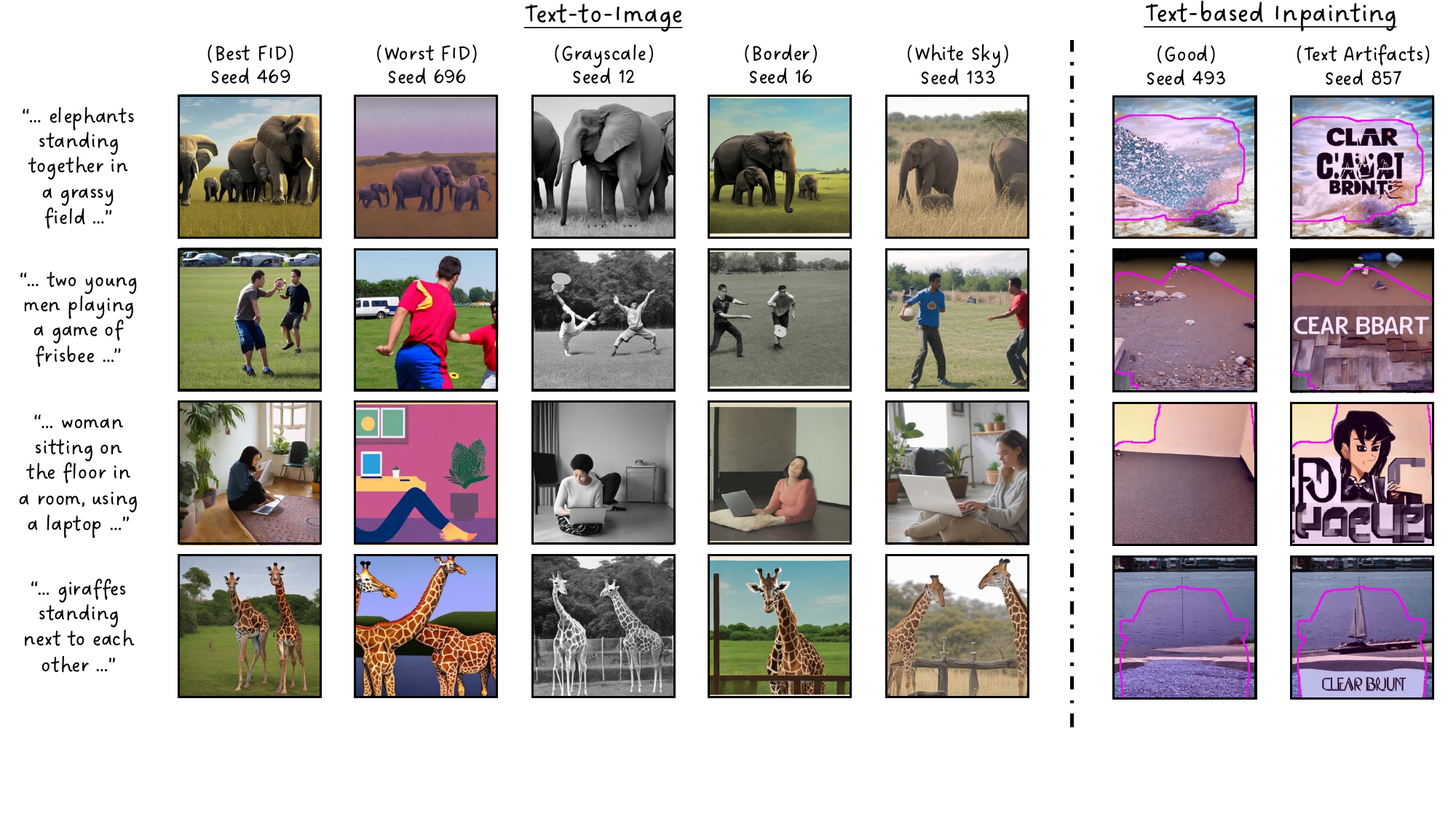}
    \vspace{-20 pt}
    \captionof{figure}[Teaser]{\textbf{Left:} Our study reveals that the seed number influences various visual elements in text-to-image generation, such as image quality and style. \textbf{Right:} Certain seeds result in more inserted text in text-based inpainting tasks like object removal.}
    \label{fig:teaser}
    \vspace{-6 pt}
\end{center}%
}]

\begin{abstract}
\vspace{-18pt}

Recent advances in text-to-image (T2I) diffusion models have facilitated creative and photorealistic image synthesis. By varying the random seeds, we can generate many images for a fixed text prompt. Technically, the seed controls the initial noise and, in multi-step diffusion inference, the noise used for reparameterization at intermediate timesteps in the reverse diffusion process. However, the specific impact of the random seed on the generated images remains relatively unexplored. In this work, we conduct a large-scale scientific study into the impact of random seeds during diffusion inference. Remarkably, we reveal that the best `golden' seed achieved an impressive FID of 21.60, compared to the worst `inferior' seed's FID of 31.97. Additionally, a classifier can predict the seed number used to generate an image with over 99.9\% accuracy in just a few epochs, establishing that seeds are highly distinguishable based on generated images. Encouraged by these findings, we examined the influence of seeds on interpretable visual dimensions. We find that certain seeds consistently produce grayscale images, prominent sky regions, or image borders. Seeds also affect image composition, including object location, size, and depth. Moreover, by leveraging these `golden' seeds, we demonstrate improved image generation such as high-fidelity inference and diversified sampling. Our investigation extends to inpainting tasks, where we uncover some seeds that tend to insert unwanted text artifacts. Overall, our extensive analyses highlight the importance of selecting good seeds and offer practical utility for image generation.
\vspace{-25pt}
\end{abstract}

\section{Introduction}
\label{sec:intro}
\vspace{-5pt}

Text-to-Image (T2I) diffusion models \cite{balaji2022ediff, betker2023improving, chen2023pixart, pernias2023wuerstchen, ramesh2022hierarchical, rombach2022high, yu2022scaling_parti} have advanced image synthesis significantly, enabling the creation of photorealistic, high-resolution images. But, their training requires substantial compute, limiting such research to a few well-equipped labs. Despite these limitations, many studies have enhanced image generation during inference by feature re-weighting \cite{si2023freeu}, gradient-based guidance \cite{epstein2023diffusion, shi2023dragdiffusion, tumanyan2023plug}, or fusion with multimodal LLMs \cite{cao2024synartifact, yang2024mastering}.

In this work, we propose an inference technique to enhance image generation by exploring `secret seeds' in the reverse diffusion process. Inspired by prior research \cite{picard2021torch}, which revealed that well-chosen neural network initialization seeds can outperform poorly chosen ones in image classification, we investigate whether `golden' or `inferior' seeds similarly impact image quality in T2I diffusion inference. Surprisingly, using the pretrained T2I model Stable Diffusion (SD) 2.0 \cite{rombach2022high} across 1,024 seeds, we discovered that the best `golden' seed achieved an FID \cite{heusel2017gans_fid, Seitzer2020FID} of \textbf{21.60}, whereas the worst `inferior' seed only reached an FID of \textbf{31.97}---a significant difference within the community. This finding sparked our curiosity to understand several scientific questions: What does the seed control in T2I diffusion inference? Why are random seeds so impactful? Can seeds be distinguished by the images they generate? Do they control interpretable image dimensions, and if so, how can this be leveraged to enhance image generation? 

To address our research questions, we first examined how random seeds control the initial noisy latent and the Gaussian noise during the reparameterization step of each intermediate timestep in the reverse latent diffusion process, as detailed in Sec. \ref{sec:seed_control}. We also developed a dataset using two T2I diffusion models: the conventional multi-step SD 2.0 \cite{rombach2022high} and the distilled one-step SDXL Turbo \cite{sauer2023adversarial}. This dataset includes over 22,000 diverse text prompts and, using 1,024 unique fixed seeds for each combination of model and prompt, resulted in approximately 46 million images as discussed in Sec. \ref{sec:data_generation}. Our initial objective was to determine whether each random seed encodes unique characteristics identifiable in the generated images. To test this, we trained a 1,024-way classifier to predict the seed number used during diffusion inference from the generated images across diverse prompts. Remarkably, this classifier reached over 99.9\% validation accuracy after just six epochs, a stark contrast to the random guessing chance of approximately 0.01\%, establishing that seeds are highly distinguishable based on the generated images as shown in Sec. \ref{sec:seed_classifier}. 

Having confirmed seed distinguishability, we aim to understand if there are interpretable perceptual dimensions enabling this differentiation. Our next step involves designing a pipeline to extract style and layout representations, apply dimensionality reduction \cite{abdi2010principal, van2008visualizing} for visible clustering, and identify consistent patterns across seeds, regardless of the input prompts. For example, certain seeds consistently produce `grayscale' images, others generate images with prominent white `sky' regions at the top, and some seeds create image borders or insert `text' during inpainting mode. In terms of image layout, various seeds consistently influence the main subject's scale, location, and depth within images. The details on these findings are in Sec. \ref{sec:impact_of_seeds}.

Building on these discoveries from our seed analysis, we propose several downstream applications to enhance image generation, as detailed in Sec. \ref{sec:applications}.
First, by identifying `golden' seeds across a variety of prompts, we can limit sampling to the top-K seeds for high-fidelity inference. This approach demonstrates superior quantitative performance, as measured by FID \cite{Seitzer2020FID} and HPS v2 \cite{wu2023human_hpsv2}, compared to random sampling in the default implementation.
Second, our findings indicate that certain seeds capture distinct styles or layout compositions. By leveraging this knowledge, we can implement diversified sampling based on style or layout, offering users varied results. 
Lastly, our studies on image inpainting reveal that some seeds consistently generate `text artifacts' instead of completing pixels, indicating that one could improve inpainting quality by using seeds that minimize these artifacts.
Note that for all these applications, we only need to perform the seed analysis once per model, and our approach can be easily integrated into the inference process \emph{without adding any computational overhead, unlike most optimization-based approaches}.

We summarize our contributions as follows:
\begin{itemize}[noitemsep, nolistsep]
  \item We present the first large-scale seed analysis for text-to-image diffusion models and have constructed a dataset comprising over 46 million images generated from both a multi-step and a one-step diffusion models, across a diverse range of text prompts.
  \item We discovered that seeds encode highly discriminative information, enabling a classifier to predict the source seed from 1,024 possible seeds with 99.9\% validation accuracy using only the generated image as input.
  \item We found that seeds significantly influence image quality, style, layout composition, and the generation of `text artifacts' across various prompts.
  \item Capitalizing on our insights, we propose applications that enhance high-fidelity or diversified inference for T2I models, and improve generation quality by avoiding `text artifacts' in text-based inpainting models.
\end{itemize}

\vspace{-5pt}
\section{Related Work}
\label{sec:related}
\vspace{-5pt}

\textbf{Stochasticity in deep learning models.}
Prior works \cite{bouthillier2021accounting, jordan2023calibrated, mehrer2020individual, picard2021torch, scabini2022improving} have primarily examined the stochasticity in neural network training caused by randomly initialized weights, random data ordering, and stochastic optimization. Notably, Picard \cite{picard2021torch} identified a significant difference of 1.82\% in test accuracy on CIFAR-10 \cite{cifar10} between the best and worst seeds, highlighting the considerable impact of the seed on model performance. Inspired by these findings, we explore the randomness within the reverse diffusion process of T2I diffusion models.

\begin{figure*}[!th]
 \vspace{-24 pt}
    \centering
    \includegraphics[trim=0in 4in 0in 0in, clip,width=\textwidth]{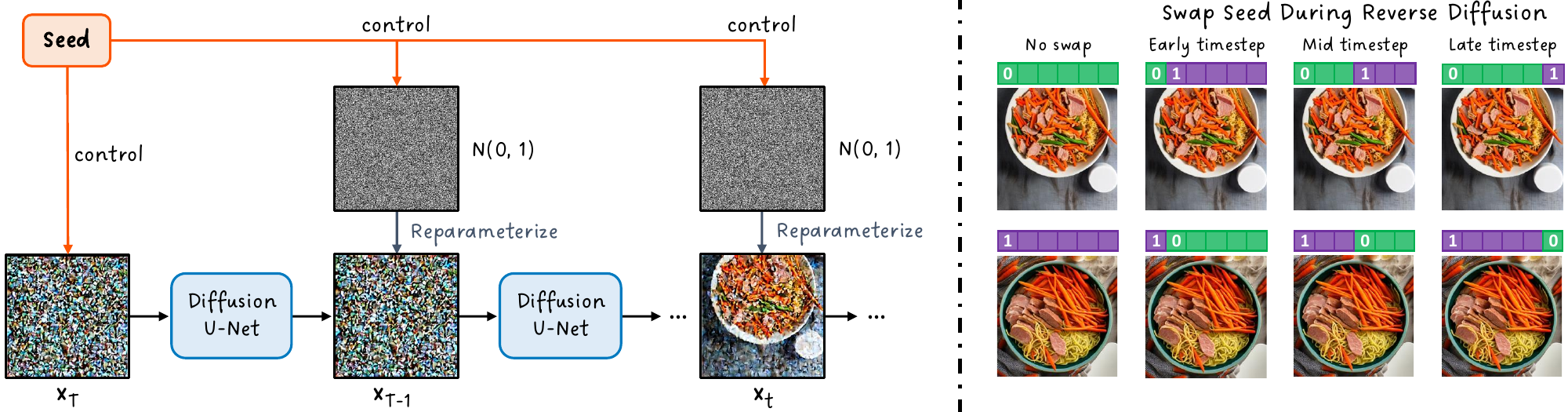}
    \vspace{-20 pt}
    \caption{\textbf{Left:} Overview of how the seed controls the initial noise $x_T$ and intermediate $x_t$ via the sampled noise in multi-step diffusion inference. \textbf{Right:} We swap the seed number at early, mid, and late timesteps of the reverse diffusion process, showing an example with seeds \textcolor{green}{0} and \textcolor{purple}{1}. Interestingly, the seed mostly influences the initial noisy latent, rather than intermediate timesteps.}
    \label{fig:what_seed_controls}
    \vspace{-15 pt}
\end{figure*}

\textbf{Impact of diffusion model inputs.}
The main sources of variation in images produced by pretrained text-to-image diffusion models \cite{balaji2022ediff, betker2023improving, chen2023pixart, pernias2023wuerstchen, ramesh2022hierarchical, rombach2022high, yu2022scaling_parti} are the text prompt and the random seed that controls the initial noise. Consequently, carefully selecting these model inputs can enhance image generation and editing during inference without requiring additional model training or fine-tuning. Several studies \cite{patashnik2023localizing, toker2024diffusion_lens, wu2023uncovering_disentanglement, yu2024uncovering} have focused on understanding the impact of text embeddings on the generated image or leveraging these text embeddings for tuning-free image generation. Yu et al. \cite{yu2024uncovering} discovered that the CLIP \cite{radford2021learning_clip} text embedding commonly used in T2I diffusion models contains diverse semantic directions that facilitate controllable image editing.
Furthermore, recent works \cite{guo2024initno, mao2023guided, mao2023semantic_winning_tickets, po2023synthetic_shifts} have shown that the initial noise can lead to certain image generation tendencies. Po-Yuan et al. \cite{po2023synthetic_shifts} demonstrated that slight perturbations to the initial noise can greatly alter the generated samples of a diffusion model, and Grimal et al. studied the effect of seeds on text-image alignment \cite{grimal2024tiam}. However, the extent to which the initial noise affects various visual dimensions of the image remains unclear. Therefore, we conduct an extensive analysis of the influence of random seeds on the generated image's quality, human preference alignment, style, composition, and insertion of `text artifacts.'

\textbf{Optimizing initial noise in diffusion models.}
Given the significant impact of the seed on images from T2I diffusion models, previous works \cite{chen2024tino, guo2024initno, mao2023guided, mao2023semantic_winning_tickets, samuel2023all_seedselect} have aimed to optimize the initial noise to produce images that better align with a text prompt, reduce visual artifacts, or achieve a desired layout. Mao et al. \cite{mao2023semantic_winning_tickets} found that certain patches of initial noise are more likely to denoise into specific concepts, enabling them to approach image editing by simply substituting regions of the initial noise without fine-tuning or disrupting the reverse diffusion process. While their work focuses on a local analysis of the initial noise, our research provides a large-scale study of the random seeds that control the initial noise across a diverse set of text prompts.

\vspace{-7pt}
\section{Understanding Diffusion Seeds}
\label{sec:theory}
\vspace{-4pt}

\begin{figure*}[!h]
 \vspace{-24 pt}
    \centering
    \includegraphics[trim=0in 5.2in 3.6in 0in, clip,width=\textwidth]{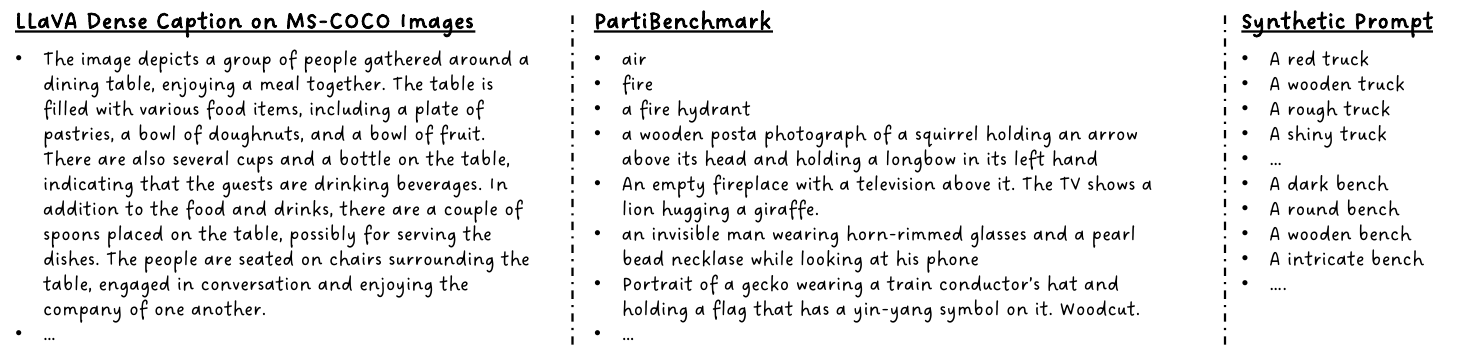}
    \vspace{-20 pt}
    \caption{A visualization of three different types of text prompts used in our study.}
    \label{fig:prompts}
    \vspace{-15 pt}
\end{figure*}

\subsection{What do seeds control in the reverse diffusion process?}
\label{sec:seed_control}
\vspace{-5pt}

Random seeds play different roles in deep learning depending on the context. During deep network training, they often influence the initialization of neural network weights, data scheduling, augmentation strategies, and stochastic regularization techniques such as dropout \cite{srivastava2014dropout}. In this work, we aim to understand what the seeds control in the reverse diffusion process and during diffusion inference.

We focus on latent diffusion models as described by Rombach et al. \cite{rombach2022high}, but the same principles apply to pixel diffusion models. Theoretically, in the traditional multi-step reverse diffusion process, both the initial noisy latent variables and the noise used for reparameterization \cite{kingma2013auto} at each timestep are sampled from a Gaussian distribution, introducing randomness. We show this process on the left side of Fig. \ref{fig:what_seed_controls}. At the implementation level, we confirmed that random seeds are inputs to compute these variables \cite{von-platen-etal-2022-diffusers}. In a distilled one-step diffusion model, such as SDXL Turbo \cite{sauer2023adversarial}, the random seeds only determine the initial noisy latent, as there are no intermediate denoising steps.

In multi-step diffusion inference, seeds determine both the initial latent variables and the reparameterization noise at each timestep. To understand the separate impacts of the initial latent configuration and the reparameterization step on the generated images, we conducted a simple "seed swap" study shown on the right side of Fig. \ref{fig:what_seed_controls} using the DDIM scheduler \cite{song2020denoising_ddim} with 40 inference steps. In our study, we first set the seed to $i$ and begin the reverse diffusion process. Then, at an intermediate timestep, we change the seed to $j$ and complete the image generation process. We explore using seeds 0 and 1 for both $i$ and $j$, as well as swapping the seed at early, mid, and late timesteps of the reverse diffusion process. Despite these variations, we found that the initial noisy latent significantly controls the generated content, while the random noise introduced at intermediate reparameterization steps has no visible impact on the generated images, as shown on the right side of Fig. \ref{fig:what_seed_controls}.

\vspace{-3pt}
\subsection{Data Generation}
\label{sec:data_generation}
\vspace{-3pt}

To conduct seed analysis at large scale, we employ three types of prompts for text-to-image (T2I) generation, as shown in Fig. \ref{fig:prompts}. First, we capture a broad spectrum of natural visual content by sampling 20,000 images from the MS-COCO 2017 train set \cite{lin2014microsoft} and generate dense captions using LLaVA 1.5 \cite{liu2023improved}. Second, we utilize 1,632 prompts from the PartiPrompts benchmark \cite{yu2022scaling_parti}, which includes short and long general-purpose user prompts. Lastly, to enable more controlled scientific studies, we create synthetic prompts by combining 40 object categories with 22 modifiers, resulting in 880 unique combinations.

For each prompt in our dataset, we sample 1,024 seeds ranging from 0 to 1,023 and generate images using two T2I models, SD 2.0 \cite{rombach2022high} and SDXL Turbo \cite{sauer2023adversarial}, for a large-scale seed analysis. Specifically, we assign the seed via torch.Generator(``cuda").manual\_seed(seed) and use a DDIM scheduler \cite{song2020denoising_ddim} for SD 2.0.
This results in a total number of $22,512$ prompts $ \times 1,024 $ seeds $ \times 2 $ models $ = 46,104,576$ images. Beyond text-to-image applications, we curated 500 pairs of images and masks for diffusion inpainting models based on Open Images \cite{OpenImages, OpenImages2}, where the mask typically covers an object in the original image. For the text prompts, we use ``clear background" to simulate the object removal case and the original object category for the object completion case, where the details are in Sec. \ref{sec:improved_inpainting}.

\vspace{-4pt}
\subsection{How discriminative are seeds based on their generated images?}
\label{sec:seed_classifier}
\vspace{-4pt}

\begin{figure}[h!]
 \vspace{-6 pt}
    \centering
    \includegraphics[trim=0in 4.5in 5in 0in, clip,width=0.95\columnwidth]{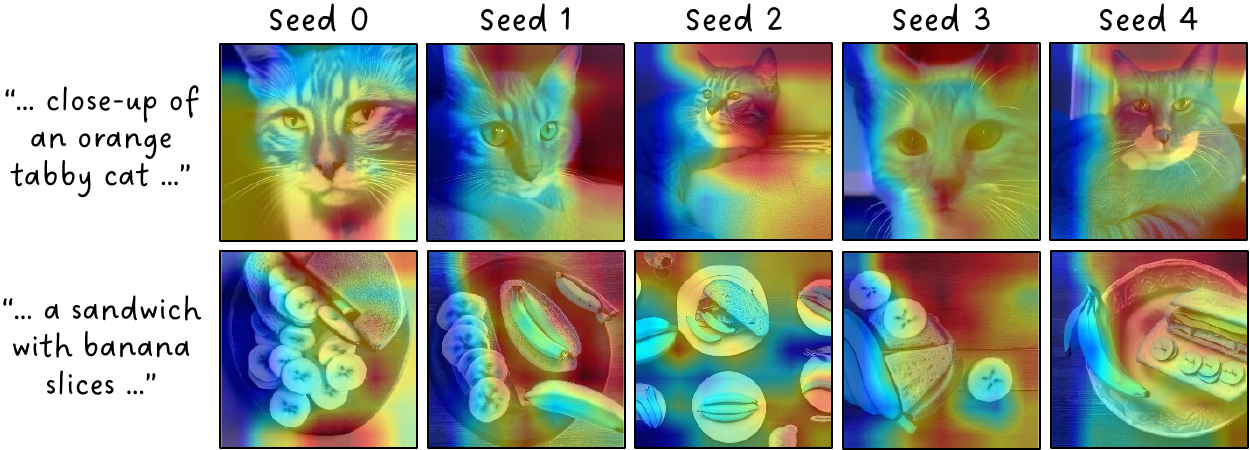}
    \vspace{-6 pt}
    \caption{Grad-CAM \cite{jacobgilpytorchcam, selvaraju2017gradcam} of our classifier trained to predict the seed used to create an image.}
    \label{fig:gradcam}
    \vspace{-14 pt}
\end{figure}

As an initial experiment, we examine whether seeds can be distinguished by their generated images. We train a 1,024-way classifier to predict the seed number used to produce an image, employing 9,000 training, 1,000 validation, and 1,000 test images per seed. Remarkably, seeds are highly differentiable based on their images. After only six epochs, our classifier trained on images from SD 2.0 \cite{rombach2022high} achieved a test accuracy of 99.99\%, and the classifier trained on images from SDXL Turbo \cite{sauer2023adversarial} reached a test accuracy of 99.96\%. However, it is unclear what makes seeds easily discernible, as the Grad-CAM \cite{jacobgilpytorchcam, selvaraju2017gradcam} visualization in Fig. \ref{fig:gradcam} is not clearly interpretable. These findings suggest that seeds may encode unique visual features; thus, we explore their impact across several interpretable dimensions.

\subsection{Impact of Seeds on Interpretable Dimensions}
\label{sec:impact_of_seeds}
\vspace{-4pt}

In Sec. \ref{sec:seed_classifier}, we observed that a classifier trained to predict the seed used to generate an image achieves over 99.9\% accuracy in just a few epochs of training. However, it remains unclear what aspects of the generated images enable these seeds to be highly distinguishable. Therefore, we present an extensive empirical study on the influence of seed number on interpretable visual dimensions.

\begin{figure*}[h]
 \vspace{-24 pt}
    \centering
    \includegraphics[trim=0in 0.75in 1.4in 0in, clip,width=\textwidth]{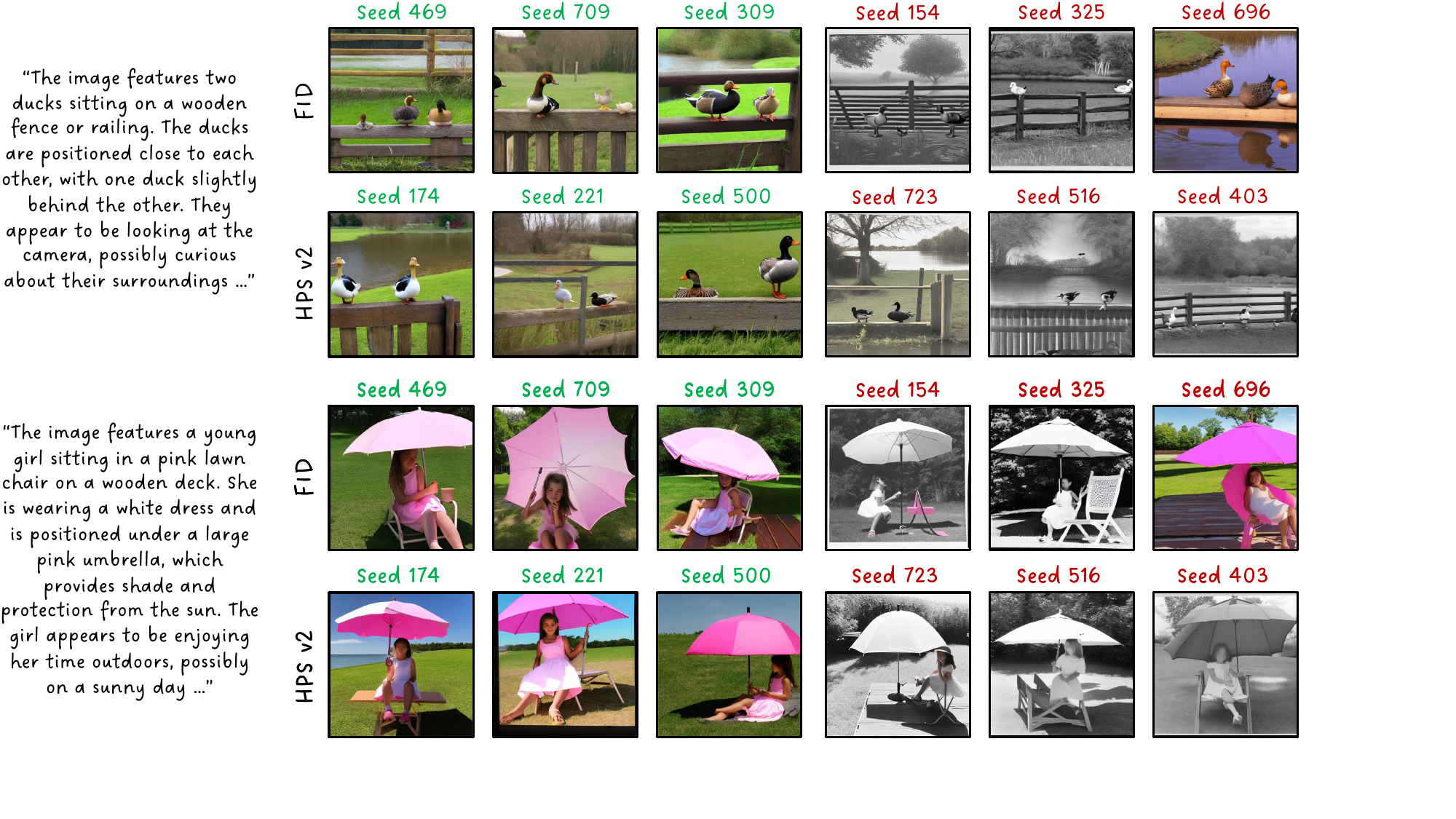}
    \vspace{-18 pt}
    \caption{We compare the top three best and worst seeds for SD 2.0 using FID \cite{heusel2017gans_fid} and HPS v2 \cite{wu2023human_hpsv2}. }
    \label{fig:sd2_good_bad_seeds}
    \vspace{-4 pt}
\end{figure*}

\begin{figure*}[!h]
    \centering
    \includegraphics[trim=0in 5.7in 0in 0in, clip,width=\textwidth]{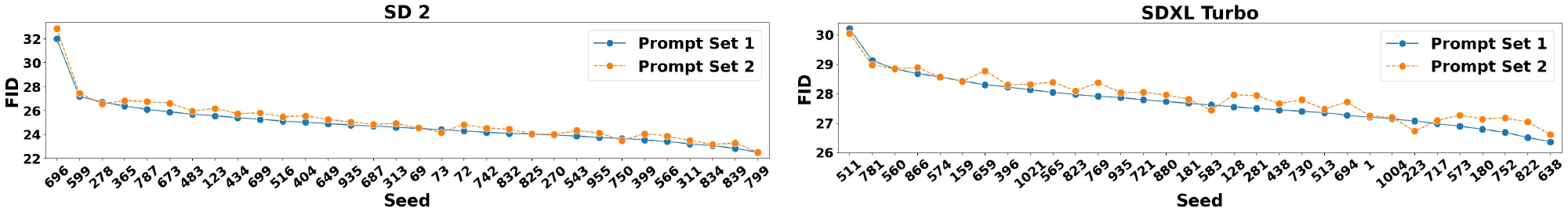}
    \vspace{-20 pt}
    \caption{We sort seeds by FID \cite{heusel2017gans_fid} using 10,000 images in Prompt Set 1, and then display the FID for the same seeds using another 10,000 images in Prompt Set 2. Lower FID indicates better quality. }
    \label{fig:t2i_fid}
    \vspace{-4 pt}
\end{figure*}

\begin{figure*}[!h]
    \centering
    \includegraphics[trim=0in 5.7in 0in 0in, clip,width=\textwidth]{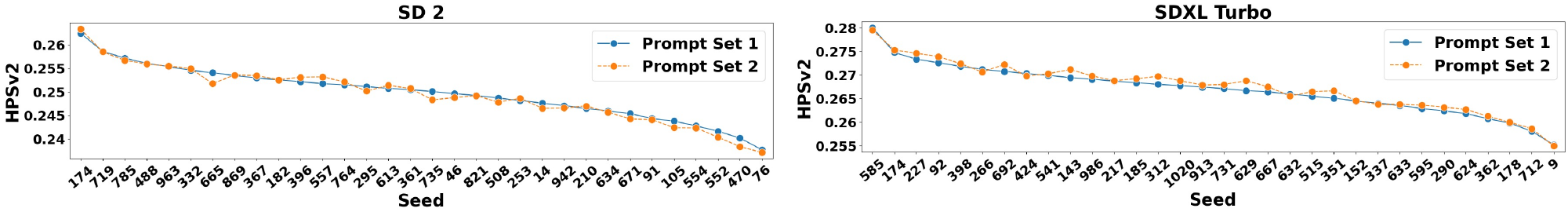}
    \vspace{-20 pt}
    \caption{We sort seeds by Human Preference Score v2 \cite{wu2023human_hpsv2} using 1,000 images in Prompt Set 1, and then plot the score for the same seeds using another 1,000 images in Prompt Set 2. A higher HPS v2 score indicates the images are more aligned with human preferences.}
    \label{fig:t2i_hpsv2}
    \vspace{-15 pt}
\end{figure*}

\textbf{Image Quality and Human Preference Alignment.}
As mentioned in Sec. \ref{sec:data_generation}, we used 20,000 prompts from MS-COCO dense captions \cite{lin2014microsoft, liu2023improved}. For each prompt, we generated images using 1,024 seeds. To evaluate the image quality associated with each seed, we selected 10,000 prompts and their corresponding generated images, and then computed the FID score \cite{heusel2017gans_fid, Seitzer2020FID} against 10,000 real MS-COCO images \cite{lin2014microsoft}. Surprisingly, we observed a significant difference in FID scores between the best and worst seeds. For instance, the `golden' seed 469 for SD 2.0 achieved a low FID of 21.60, while the `inferior' seed 696 scored 31.97---a disparity considered significant within the community. Additionally, we assess the seeds using HPS v2 \cite{wu2023human_hpsv2}, a new metric trained on large-scale human preference pairs to quantify human preferences for AI-generated images. For each seed, we sampled 1,000 prompts and their corresponding images to calculate HPS v2. As shown in Fig. \ref{fig:sd2_good_bad_seeds}, the top and bottom three seeds according to FID and HPS v2 indeed reveal that the highest-rated seeds produce images that are more visually pleasing and aligned with human preferences.

Next, we determine whether these seed rankings are generalizable across a different set of 10,000 prompts for FID and 1,000 prompts for HPS v2. In Fig. \ref{fig:t2i_fid} and \ref{fig:t2i_hpsv2}, we plot the ranked seeds for FID and HPS v2 using images from SD 2.0 and SDXL Turbo. We compare scores from the first set of prompts (``Prompt Set 1") against scores from another set of prompts (``Prompt Set 2"). We reveal a high degree of overlap between the seed patterns for quality and human preference. This consistency underpins our proposed enhancements to inference strategies in Sec. \ref{sec:high_fidelity_inference} and \ref{sec:diverse_inference}.

\begin{figure*}[!th]
 \vspace{-24 pt}
    \centering
    \includegraphics[trim=0in 0.18in 0.0in 0in, clip,width=\textwidth]{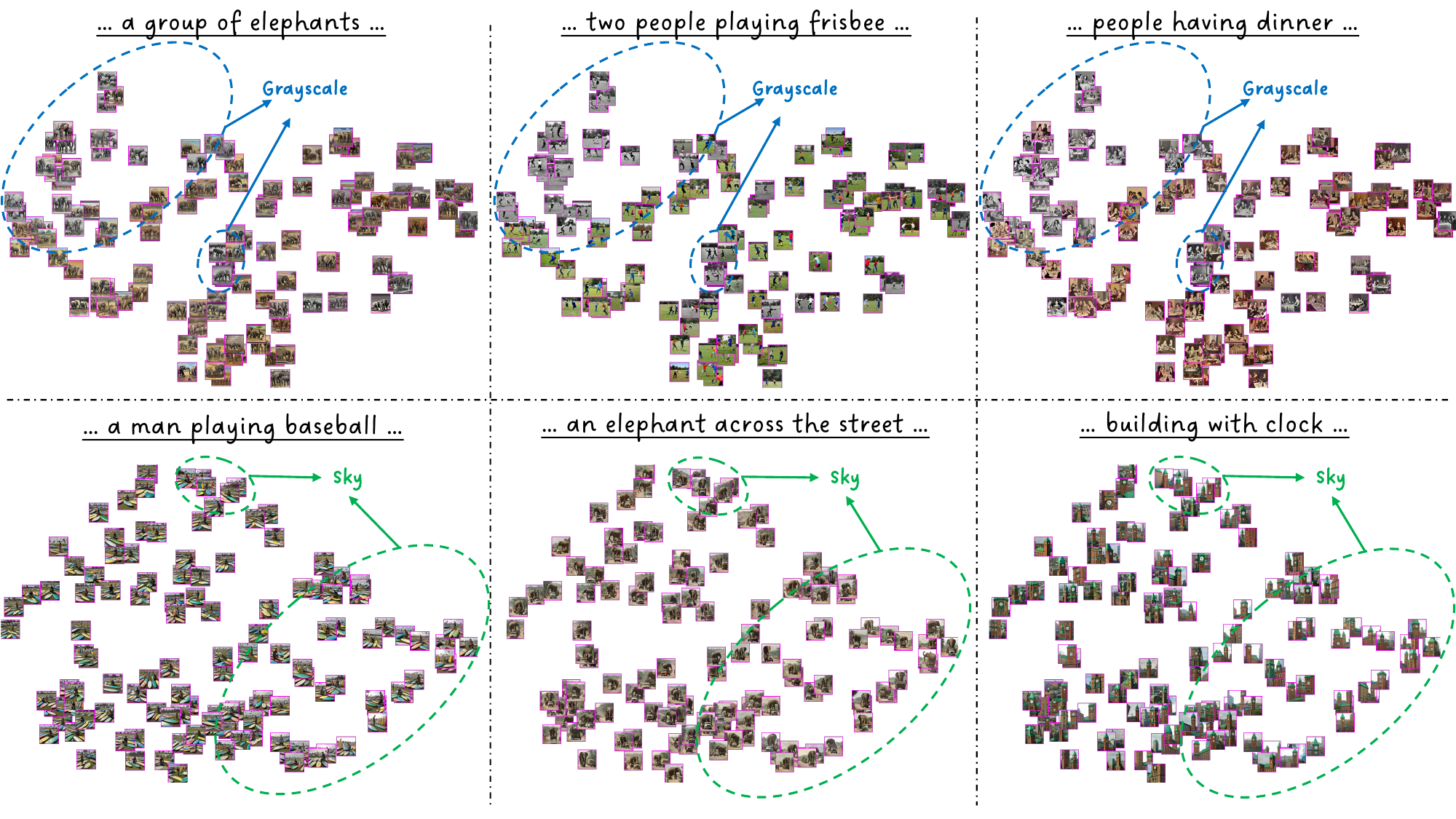}
    \vspace{-20 pt}
    \caption{Style embedding clustering across various prompts, with each position corresponding to a unique seed. Certain seeds tend to generate grayscale images for SD 2.0 (top), while others frequently produce images with `white sky' regions for SDXL Turbo (bottom). \textbf{Please zoom-in to check.} }
    \label{fig:style_clusters}
    \vspace{-4 pt}
\end{figure*}

\textbf{Image Style.}
Given the visual variations in images generated using different seeds, we investigate whether specific seeds consistently produce unique style patterns across prompts. Drawing on established methods in image texture and style transfer \cite{gatys2015texture, gatys2016image}, we compute style representations by extracting the Gram matrix — which measures pairwise cosine similarity across channels — from a pretrained deep network \cite{simonyan2014very_vgg} at multiple layers. Next, we reshape the Gram matrix into a single-column vector for each image and reduce its dimensionality to two using PCA and t-SNE \cite{abdi2010principal, van2008visualizing}. Now, for each image, we have a compact 2D vector that captures its style. Using $N = 1024$ seeds and $P$ prompts results in a feature dimension of $N \times (2 \times P)$, combining the style representation across the generated images for each seed. We further reduce \cite{abdi2010principal, van2008visualizing} this aggregated style representation per seed from $N \times (2 \times P)$ to $N \times 2$. Finally, a subset of seeds are visualized in Fig. \ref{fig:style_clusters}, providing a clear visual representation of style clustering at the seed level.

\begin{figure*}[!th]
    \centering
    \includegraphics[trim=0in 6.2in 0in 0in, clip,width=\textwidth]{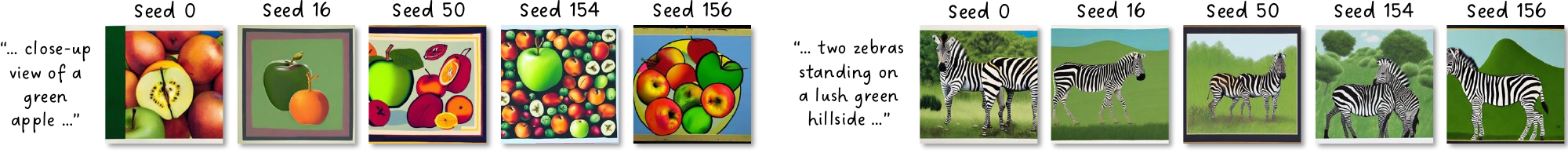}
    \vspace{-20 pt}
    \caption{Certain seeds produce a ``border" around images for SD 2.0. Often, these borders appear as horizontal bars at the top and bottom. Surprisingly, seed 0 occasionally generates a thick dark border on the left side of the image, while seed 50 sometimes adds a ``photo frame."}
    \label{fig:border_style}
    \vspace{-15 pt}
\end{figure*}

In Fig. \ref{fig:style_clusters}, the positions within the embedding space correspond to the same seeds across various subplots. As depicted in the first row, certain seed groups consistently generate grayscale images irrespective of the prompt. Similarly, the second row shows that some seeds tend to produce images with prominent sky regions, while others do not. Furthermore, in Fig. \ref{fig:border_style}, we observe that a select group of seeds consistently generate images with a `border' effect near the edges, regardless of the text prompt. Collectively, these findings demonstrate that individual seeds exhibit distinct tendencies in style generation across varying prompts. 

\begin{figure*}[t]
 \vspace{-24 pt}
    \centering
    \includegraphics[trim=0in 1.6in 0in 0in, clip,width=\textwidth]{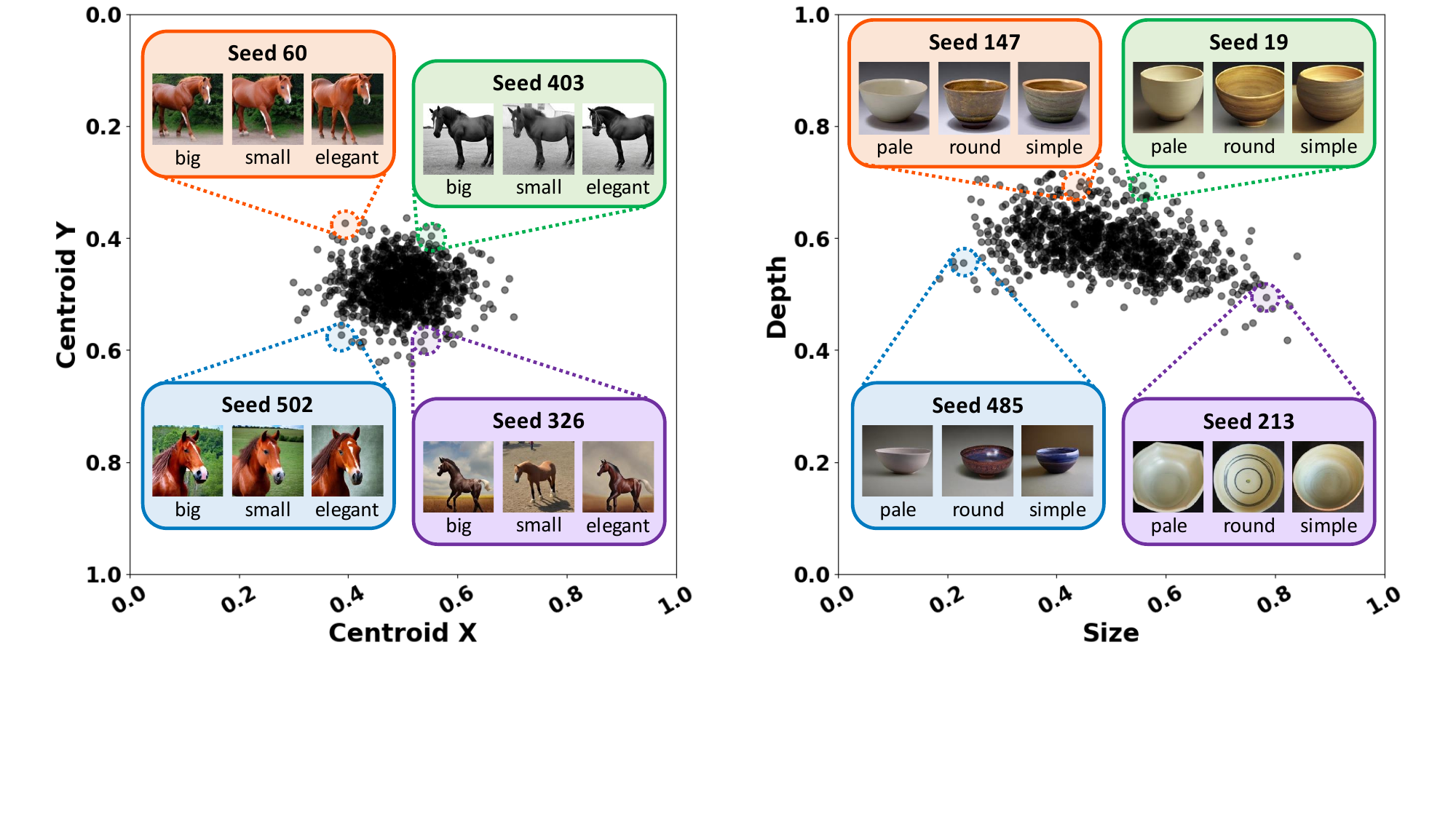}
    \vspace{-17pt}
    \caption{We observe that seeds produce images with unique and consistent compositions for a given object category. Each data point represents a seed. For each seed, we combine image composition features from 22 prompts with slight variations like ``a pale bowl" and ``a round bowl." Then, we apply dimensionality reduction \cite{abdi2010principal, van2008visualizing} for visualization. \textbf{Left:} Distribution of object centroid $(x, y)$ coordinates. \textbf{Right:} Distribution of object depth and size relative to the image.}
    \label{fig:layout}
    \vspace{-4 pt}
\end{figure*}

\textbf{Image Composition.}
Moving beyond style, we examine whether seeds create distinctive image compositions, such as consistent object locations and sizes. As described in Sec. \ref{sec:data_generation}, we generate images using 880 synthetic prompts consisting of 40 object categories paired with 22 modifiers, which includes adjectives and the empty string. For each image, we segment \cite{cheng2022mask2former} the object and compute an image composition feature vector that contains the object's centroid $(x,y)$ coordinates, size, and depth \cite{depthanything} relative to the image.
On the left side of Fig. \ref{fig:layout}, we visualize the distribution of the object mask's centroid for the category ``horse." Remarkably, the object's position stays relatively the same despite slight prompt alterations. On the right side of Fig. \ref{fig:layout}, we observe an analogous pattern in the object's size and depth for the category ``bowl." Overall, we observe that the location, size, and depth of objects are largely dependent on the specific seed used, consistent across the same object categories and irrespective of the text modifiers in the prompts.

\vspace{-5pt}
\section{Practical Applications}
\label{sec:applications}
\vspace{-2pt}

\subsection{High-Fidelity Inference}
\label{sec:high_fidelity_inference}
\vspace{-3pt}

In Sec. \ref{sec:impact_of_seeds}, we observed that `golden' seeds tend to generate images with significantly better quality and human preference alignment. This inspires us to think---how much can we improve the image quality compared to random generations by simply leveraging these `golden' seeds?

Specifically, we identified \( k \) `golden' seeds that excel in both image quality and human preference alignment. We subsequently tested these \( k \) `golden' seeds by generating images with a different set of 10,000 prompts to evaluate their performance relative to random seeds. We identified \( k=65 \) `golden' seeds for SD 2.0 and \( k=67 \) for SDXL Turbo, where \( k \) was determined by selecting seeds that ranked among the top 256 in both FID~\cite{Seitzer2020FID} and HPS v2~\cite{wu2023human_hpsv2}. We propose that a sampling pool of 60+ `golden' seeds is sufficiently large in practical applications for a single prompt. As demonstrated in Tab.~\ref{tab:high_fidelity_inference}, leveraging these well-chosen seeds significantly improves the FID and HPS v2 scores for both SD 2.0 and SDXL Turbo, and on both MS-COCO \cite{lin2014microsoft} and the PartiPrompts benchmark \cite{yu2022scaling_parti}.

\begin{table*}[h!]
\caption{We demonstrate that well-chosen seeds can outperform random generations by comparing the visual quality and human preference alignment using our `golden' seeds and random seeds. Additionally, our `golden' seeds lead to improved human preference alignment on a greater variety of prompts in the PartiPrompts benchmark \cite{yu2022scaling_parti}. Mean and standard deviation based on three trials.}
\label{tab:high_fidelity_inference}
\vspace{-8 pt}
\centering
\resizebox{\textwidth}{!}{%
\begin{tabular}{lcccccc}
\toprule
\multicolumn{1}{c}{} & \multicolumn{3}{c}{SD 2.0} & \multicolumn{3}{c}{SDXL Turbo}                  \\
\cmidrule(r){2-4} \cmidrule(l){5-7}
    & FID $(\downarrow)$ & HPS v2 $(\uparrow)$ & Parti HPS v2 $(\uparrow)$ & FID $(\downarrow)$ & HPS v2 $(\uparrow)$ & Parti HPS v2 $(\uparrow)$ \\
\midrule
Random Seeds     & 19.334 $\pm$ 0.212 & 0.250 $\pm$ 0.000 & 0.263 $\pm$ 0.001 & 24.859 $\pm$ 0.123 & 0.266 $\pm$ 0.000 & 0.290 $\pm$ 0.000 \\
Our Golden Seeds & \textbf{19.045 $\pm$ 0.058} & \textbf{0.257 $\pm$ 0.000} & \textbf{0.268} $\pm$ 0.001 & \textbf{24.209 $\pm$ 0.108} & \textbf{0.272 $\pm$ 0.000} & \textbf{0.293 $\pm$ 0.001} \\
\bottomrule
\end{tabular}
}%
\vspace{-12 pt}
\end{table*}

\subsection{Controlling Diversity in Style and Composition}
\label{sec:diverse_inference}

A typical image generation interface presents the user with four samples per prompt. Moreover, prior methods aim to promote the diversity of generated images using primarily gradient-based methods, such as Particle Guidance \cite{corso2023particle}. In Sec. \ref{sec:impact_of_seeds}, our results highlight that the choice of seed has a strong influence on the stylistic and spatial attributes in the generated images. Therefore, we explore whether we can obtain more diverse images in style or composition by merely sampling `diverse' seeds.

To select four diverse seeds per prompt, we represent each seed by a feature vector capturing its style or composition, as discussed in Sec. \ref{sec:impact_of_seeds}. We then employ farthest point sampling using these features. Specifically, we randomly pick the first seed $s_0 \sim \mathcal{U}\{0, 1023\}$ and iteratively select the next three seeds to maximize the distance in feature space from the already selected seeds.

\begin{figure*}[t]
 \vspace{-24 pt}
    \centering
    \includegraphics[trim=0in 5.18in 0in 0in, clip,width=\textwidth]{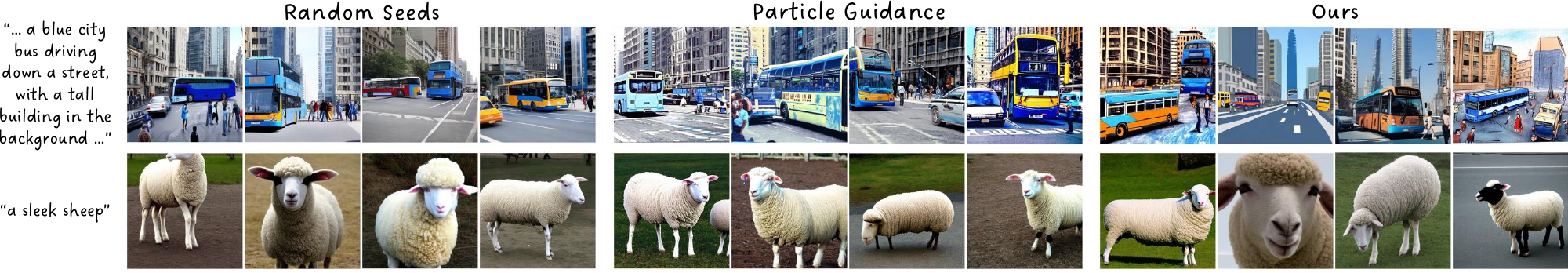}
    \vspace{-20 pt}
    \caption{We show that simply generating images using ``diverse" seeds can promote more variation in style (top) and image composition, as measured by object centroid, size, and depth (bottom).}
    \label{fig:diverse_style_layout_samples}
    \vspace{-5 pt}
\end{figure*}

\vspace{-12pt}
\begin{equation}
    s_{i} = \arg\max_{s \notin S} \min_{s' \in S} \| \mathbf{f}(s) - \mathbf{f}(s') \|, \quad \text{for } i = 1, \ldots, C-1
\end{equation}
where $S$ is our set of diverse seeds.
To evaluate whether our well-chosen seeds improve diversity over random seeds and Particle Guidance \cite{corso2023particle}, we calculate the similarity between images synthesized from a different set of $P$ prompts, where $P=500$ LLaVA \cite{liu2023improved} dense captions for image style and $P=440$ synthetic prompts for image composition. Specifically, we measure the pairwise cosine similarity of image features and average the similarity scores across prompts. A lower pairwise similarity score means higher diversity. Mathematically, the similarity is computed as:

\vspace{-16pt}
\begin{equation}
\text{Similarity} = \frac{1}{P} \sum_{i=1}^{P} \left( \frac{1}{\binom{C}{2}} \sum_{j=1}^{C} \sum_{k=j+1}^{C} \cos(\mathbf{f}_{ij}, \mathbf{f}_{ik}) \right)
\end{equation}
\vspace{-12pt}

where there are $P$ prompts and $\mathbf{f}$ denotes the feature vector representing image style or composition. We typically use $C=4$ images per prompt, but it's important to note that if no objects are detected in an image, then the image is not used to compute similarity.
In Tab. \ref{tab:diverse_style_layout}, we observe that our diverse seeds outperform random seeds and Particle Guidance \cite{corso2023particle} in generating images with varying styles and compositions for SD 2.0. Interestingly, our well-chosen seeds aid in diversifying image composition for SD 2.0 but not for SDXL Turbo. We show comparisons in Fig. \ref{fig:diverse_style_layout_samples}.

\begin{table}[h!]
\caption{We compare the style and composition diversity of images generated using our diverse seeds, Particle Guidance \cite{corso2023particle}, and random seeds. More diverse generations have lower similarity scores. We show the mean and standard deviation based on three trials.}
\label{tab:diverse_style_layout}
\vspace{-8pt}
\centering
\resizebox{\columnwidth}{!}{%
\begin{tabular}{lcccc}
\toprule
\multicolumn{1}{c}{} & \multicolumn{2}{c}{SD 2.0} & \multicolumn{2}{c}{SDXL Turbo}   \\
\cmidrule(r){2-3} \cmidrule(l){4-5}
    & Style $(\downarrow)$ & Composition $(\downarrow)$ & Style $(\downarrow)$ & Composition $(\downarrow)$ \\
\midrule
Random Seeds      & 0.98 $\pm$ 0.00  & 0.97 $\pm$ 0.00  & 0.99 $\pm$ 0.00  & 0.99 $\pm$ 0.00  \\
Particle Guidance & 0.98 $\pm$ 0.00  & 0.97 $\pm$ 0.00  & --- & ---   \\
Our Diverse Seeds & \textbf{0.97 $\pm$ 0.00} & \textbf{0.96 $\pm$ 0.00} & \textbf{0.98 $\pm$ 0.00} & 0.99 $\pm$ 0.00 \\
\bottomrule
\end{tabular}
}%
\vspace{-7 pt}
\end{table}

\begin{figure*}[t]
    \centering
    \includegraphics[trim=0in 0.1in 0in 0in, clip,width=\textwidth]{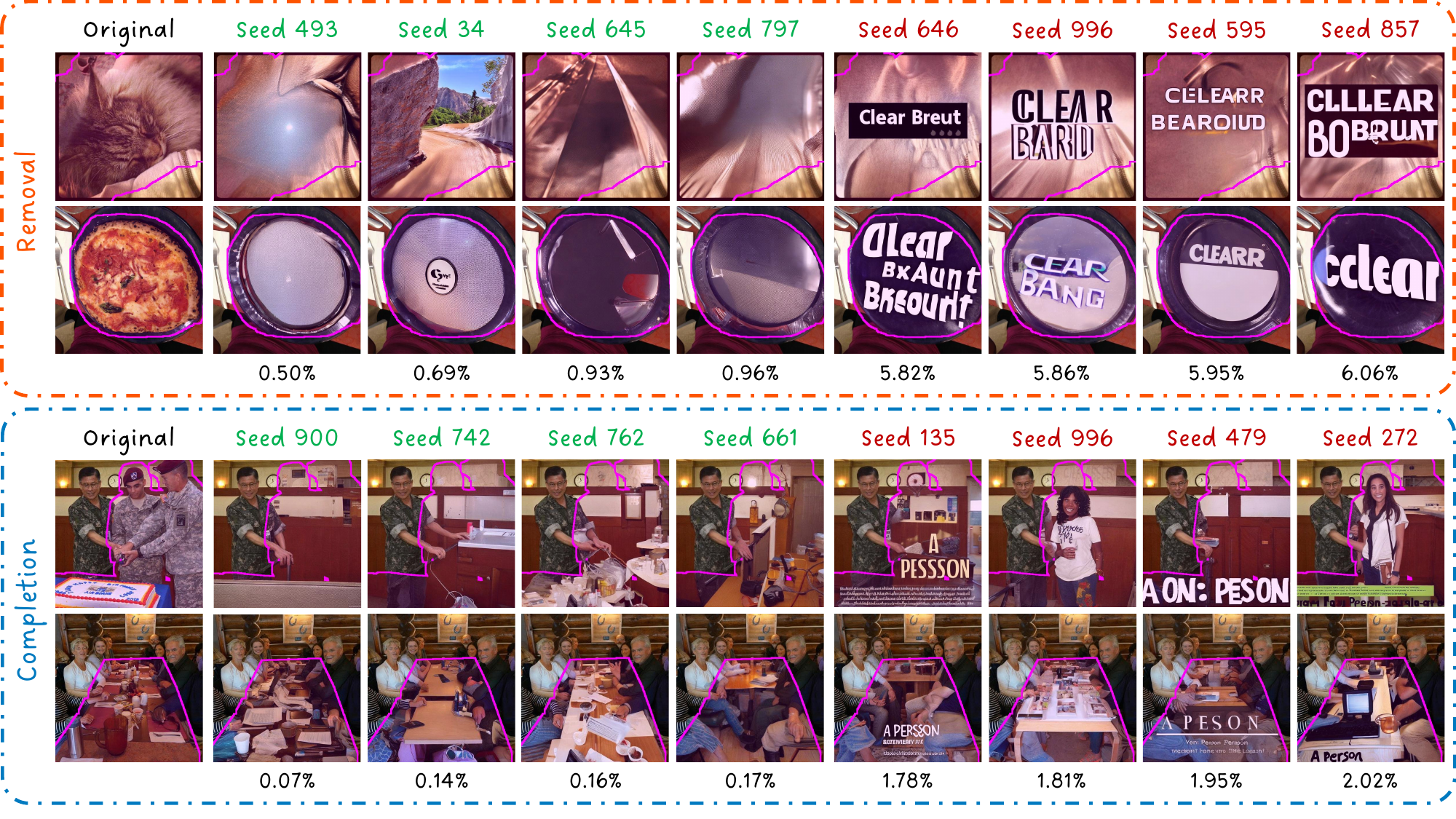}
    \vspace{-20 pt}
    \caption{Certain seeds tend to insert unwanted text within the inpainting region, outlined in pink. \textbf{Top:} We aim to remove the object using the prompt ``clear background." \textbf{Bottom:} We attempt to complete the object using a prompt that specifies the object category.}
    \label{fig:inpaint_ocr}
    \vspace{-16 pt}
\end{figure*}

\subsection{Improved Text-based Inpainting}
\label{sec:improved_inpainting}
\vspace{-3pt}

In Sec. \ref{sec:high_fidelity_inference} and \ref{sec:diverse_inference}, we showed that carefully selecting the seed provides a straightforward, training-free approach to enhance the visual quality, human preference, and diversity of images generated by text-to-image diffusion models. But, the potential of image generation extends beyond text-to-image applications. This poses an intriguing question---can we also uncover `golden' seeds for text-based image inpainting tasks, such as object removal and completion?

As described in Sec. \ref{sec:data_generation}, we gathered 500 pairs of images and inpainting masks for the removal and completion applications based on the Open Images dataset \cite{OpenImages, OpenImages2}. We employed the prompt ``clear background" for the removal case, and we used a prompt corresponding to the original object category for the completion case. We then generated images using a text-based diffusion inpainting model.
We observed that some images contain unwanted text in the inpainting region that often mimics the prompt. To quantify the presence of text, we applied optical character recognition \cite{pytesseract} and computed the mean proportion of text artifacts within the inpainting mask across all images from each seed. As shown in Fig. \ref{fig:inpaint_ocr}, certain seeds tend to insert text in both removal and completion scenarios.

\vspace{-5pt}
\section{Conclusion}
\label{sec:conclusion}
\vspace{-6pt}

In this work, we investigated the role of ``random" seeds in the reverse diffusion process, exploring their differentiability based on generated images and their impact on interpretable visual dimensions. Notably, our 1,024-way classifier trained to predict the seed number for a generated image achieved over 99.9\% test accuracy in just a few epochs. Encouraged by this finding, we conducted extensive analyses and identified `golden' seeds that consistently produce images with better visual quality and human preference alignment. Additionally, we discovered that certain seeds create `grayscale' images, add borders, or insert text during inpainting. Our studies also show that seeds influence the image composition, affecting object position, size, and depth. Leveraging these insights, we propose downstream applications such as high-fidelity inference and diversified generation for text-to-image diffusion models by merely sampling these special seeds. Our analyses offer new perspectives on enhancing image synthesis during inference without significant computational overhead.

\textbf{Acknowledgements.}
This work is supported by funds provided by the National Science Foundation and by DoD OUSD (R\&E) under Cooperative Agreement PHY-2229929 (The NSF AI Institute for Artificial and Natural Intelligence).

{\small
\bibliographystyle{ieee_fullname}
\bibliography{egbib}
}

\newpage
\onecolumn
\appendix
\twocolumn

\section{Data Generation}
\label{sec:supp_data_generation}

In Sec. 3.2, we employed pretrained model checkpoints and implementations from the Hugging Face diffusers library \cite{von-platen-etal-2022-diffusers}. Specifically, for text-to-image generation, we used Stable Diffusion 2.0 (``stabilityai/stable-diffusion-2-base") with a DDIM scheduler, and SDXL Turbo (``stabilityai/sdxl-turbo"). For text-based image inpainting, we utilized the SD 2.0 inpainting model (``stabilityai/stable-diffusion-2-inpainting"). Furthermore, our 1,024 seeds range from 0 to 1,023 inclusive, and we use \texttt{torch.Generator("cuda").manual\_seed(seed)} to assign the seed used by the model.

\subsection{Synthetic Prompts for Image Composition Analysis}
\label{sec:supp_synthetic_prompts}
We create a set of 880 prompts by pairing 40 object categories with 22 modifiers in the format ``a [modifier] [object category]". These modifiers include 21 adjectives and the empty string.
\begin{itemize}
    \item \textbf{Adjectives:} big, small, red, blue, pale, dark, transparent, shiny, dull, rustic, smooth, rough, bright, muted, round, simple, elegant, antique, monochrome, intricate, sleek
    \item \textbf{Object categories:} bicycle, car, motorcycle, airplane, bus, truck, boat, fire hydrant, bench, bird, cat, dog, horse, sheep, cow, elephant, zebra, giraffe, backpack, umbrella, suitcase, sports ball, skateboard, surfboard, tennis racket, fork, knife, spoon, bowl, apple, pizza, donut, cake, chair, couch, laptop, cell phone, clock, vase, teddy bear
\end{itemize}

\subsection{Dataset for Inpainting Applications}
\label{sec:inpainting_dataset}

We curated 500 pairs of images and inpainting masks for object removal and object completion applications, as described in Sec. 3.2. In particular, for the object removal use case, we employed images and annotations from the Open Images dataset \cite{OpenImages2, OpenImages}, and we used ``clear background" as the text prompt. To create the inpainting mask, we dilated the instance segmentation mask to ensure coverage of the object. Additionally, for the object completion use case, we sampled images from the MS-COCO dataset \cite{lin2014microsoft} and used InstaOrder \cite{lee2022instance} to determine occlusion relationships to create inpainting masks. We used the category of the object to complete as the text prompt.

\subsection{Licenses for Existing Datasets}
\label{sec:licenses}

The MS-COCO dataset \cite{lin2014microsoft} and the PartiPrompts benchmark \cite{yu2022scaling_parti} are under a CC BY 4.0 license. For the Open Images dataset \cite{OpenImages2, OpenImages}, the images are under a CC BY 2.0 license and the annotations are under a CC BY 4.0 license.

\section{Classifier for Predicting Seed Number}
\label{sec:supp_classifier}

We trained a lightweight transformer, EfficientFormer-L3 \cite{li2022efficientformer}, to predict the seed used to generate an image. For our 1,024-way classification task, we utilized 9,000 training, 1,000 validation, and 1,000 test images per seed as mentioned in Sec. 3.3. The prompts for these images are dense captions by LLaVA 1.5 \cite{liu2023improved}. Moreover, we set a batch size of 128 and train for six epochs, which obtains a model checkpoint with over 99.9\% validation and test accuracy. Our classifier uses the AdamW optimizer \cite{loshchilov2019decoupled_adamw} with learning rate 0.0002 and weight decay 0.05. We apply data augmentations during training, which include resizing each image to have a shorter edge of size $224$ using bicubic interpolation, center cropping the image to size $224 \times 224$, and randomly flipping the image horizontally with probability $0.5$. During validation and testing, we only resize and center crop the images.

\section{Compute Resources}
\label{sec:supp_compute}

To generate our dataset in Sec. 3.2, we utilized 32 A100 GPUs for roughly 24 days. Additionally, all the experiments in Sec. 3.3, 3.4, and 4 were performed on an RTX 4090 GPU with 24GB of memory. One of the longest experiments was training the classifier to predict seed number in Sec. 3.3, which took at most three days.

\section{Additional Qualitative Results}

We provide extra visualizations of the Grad-CAM from our classifier that predicts seed number in Figure \ref{fig:supp_gradcam} of the supplemental. We also show more examples of seeds that often produce a `border' around the image in Figure \ref{fig:supp_border_style} of the supplemental. Moreover, we present additional examples of good seeds and seeds that generate ``text artifacts" for object removal and completion applications in Figures \ref{fig:supp_removal} and \ref{fig:supp_completion} of the supplemental, respectively.

\begin{figure*}[!h]
 \vspace{-5 pt}
    \centering
    \includegraphics[trim=0in 1.4in 0in 0in, clip,width=\textwidth]{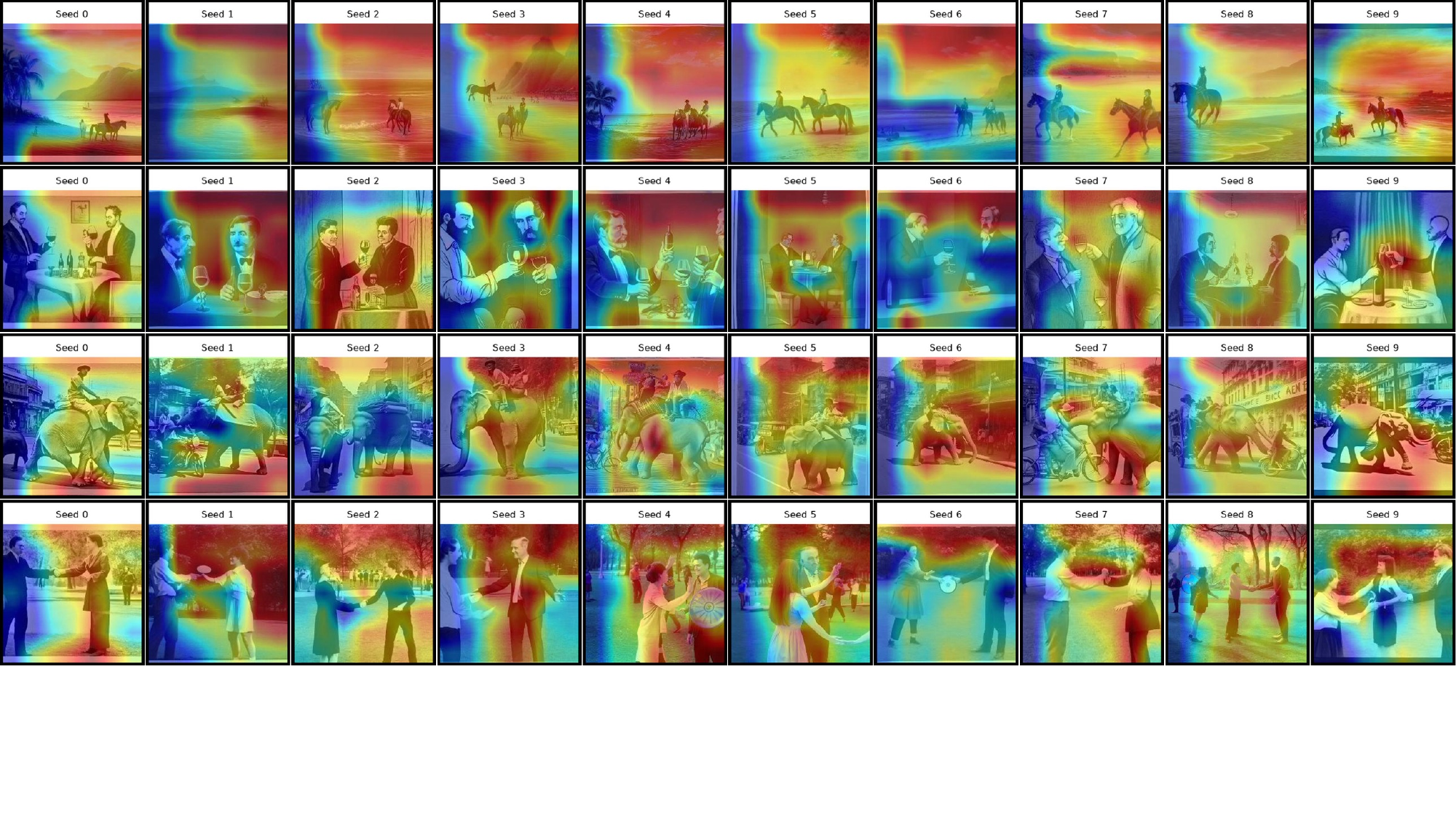}
    \vspace{-16 pt}
    \caption{Additional Grad-CAM \cite{jacobgilpytorchcam, selvaraju2017gradcam} visualizations for our classifier trained to predict the seed number for an image. We note that it is difficult to interpret what makes seeds easily distinguishable by looking at these visualizations.}
    \label{fig:supp_gradcam}
    \vspace{-4 pt}
\end{figure*}

\begin{figure*}[!h]
 \vspace{-2 pt}
    \centering
    \includegraphics[trim=0in 3.15in 0in 0in, clip,width=\textwidth]{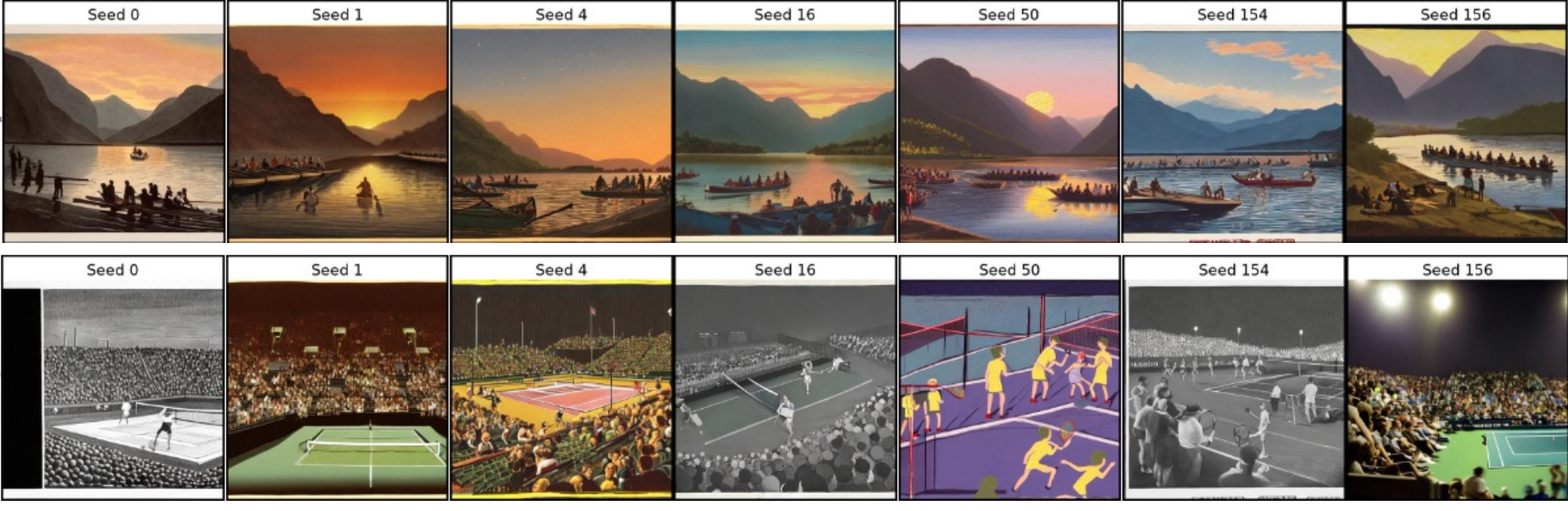}
    \vspace{-18 pt}
    \caption{Additional examples of seeds that tend to generate a `border' near the image boundaries.}
    \label{fig:supp_border_style}
    \vspace{-2 pt}
\end{figure*}

\begin{figure*}[!h]
 \vspace{-2 pt}
    \centering
    \includegraphics[trim=0in 0.6in 0in 0in, clip,width=\textwidth]{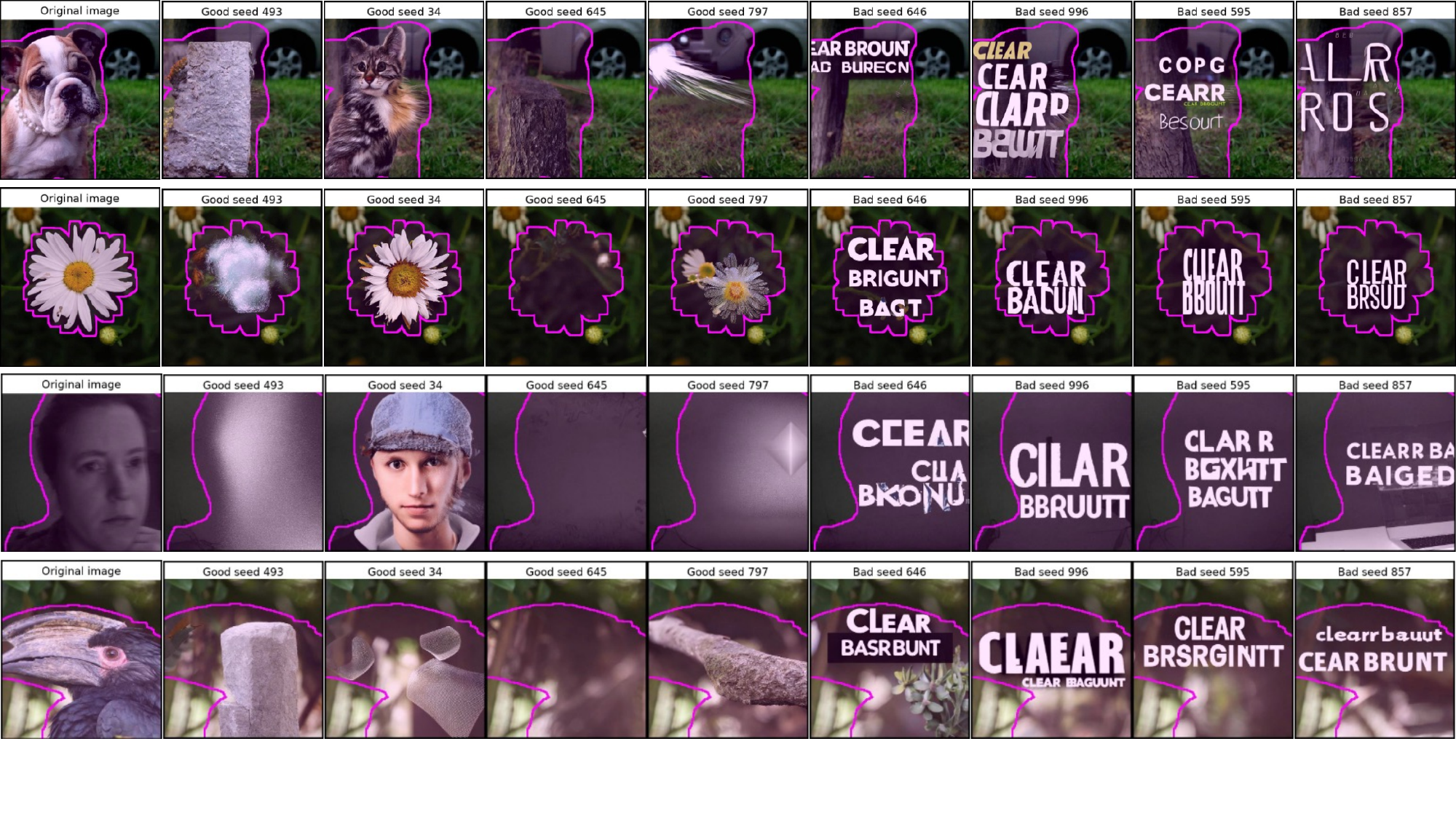}
    \vspace{-18 pt}
    \caption{Additional examples of the four best seeds and four worst seeds in terms of how much unwanted text artifacts are inserted during object removal.}
    \label{fig:supp_removal}
\end{figure*}

\begin{figure*}[!h]
 \vspace{-2 pt}
    \centering
    \includegraphics[trim=0in 0.6in 0in 0in, clip,width=\textwidth]{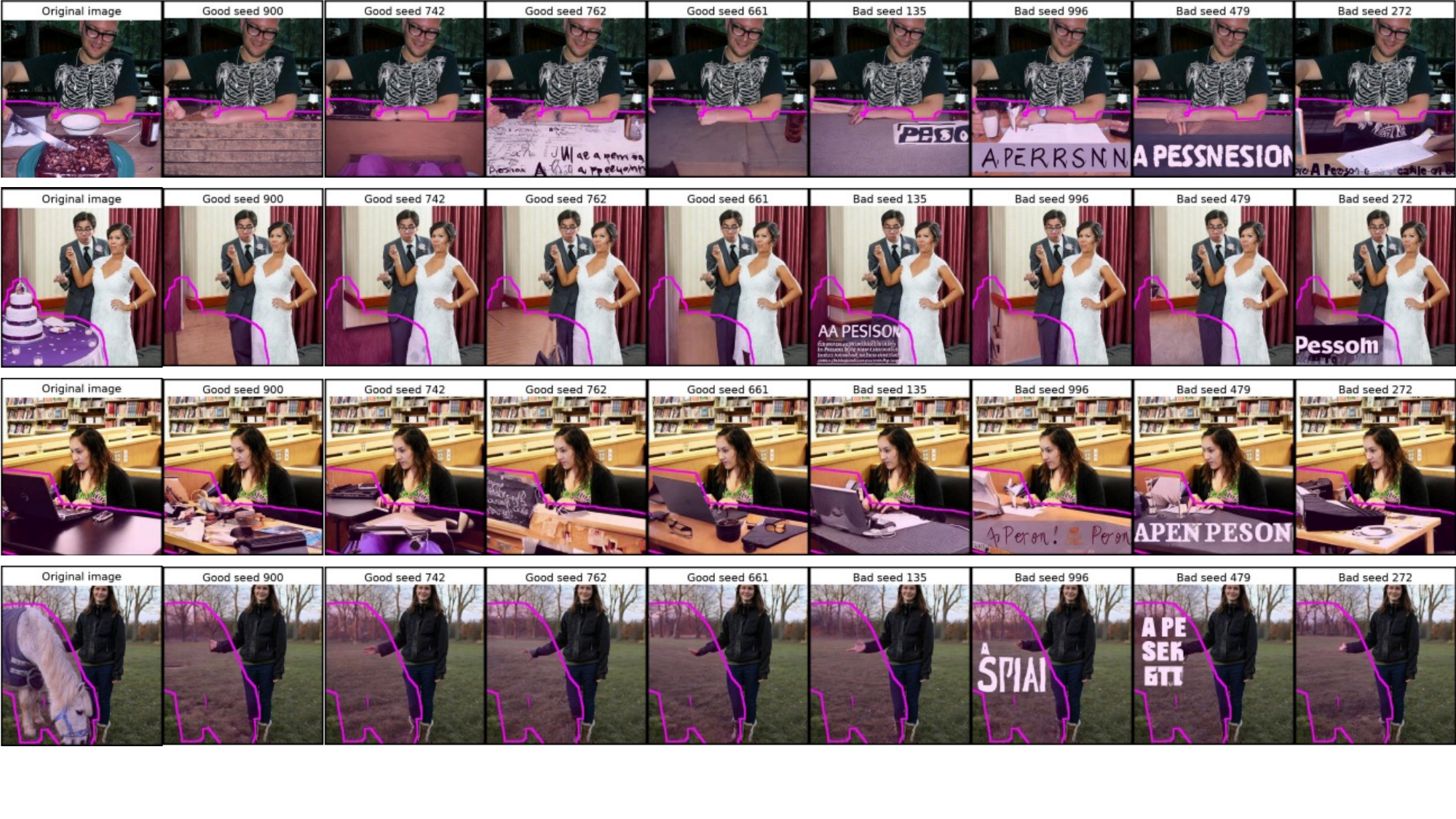}
    \vspace{-18 pt}
    \caption{Additional examples of the four best seeds and four worst seeds in terms of how much unwanted text artifacts are inserted during object completion.}
    \label{fig:supp_completion}
\end{figure*}

\end{document}